\documentclass[11pt,letterpaper]{article}
\usepackage{emnlp2017}
\usepackage{times}
\usepackage{latexsym}

\emnlpfinalcopy

\def\emnlppaperid{***}

\usepackage{url}
\usepackage{verbatim}
\usepackage{multirow}
\usepackage{caption}
\usepackage{subcaption}
\usepackage{amsmath,amssymb,amsfonts}
\usepackage[linesnumbered,ruled,vlined]{algorithm2e}
\SetKwInput{KwInput}{Input}
\SetKwInput{KwOutput}{Output}
\SetKwProg{myproc}{Procedure}{}{}

\usepackage{algpseudocode}
\usepackage{mathtools}
\usepackage{listings}

\usepackage{ifpdf}
\usepackage{lmodern}
\usepackage{lipsum}
\usepackage{textcomp}
\usepackage{enumitem}
\usepackage{mdwlist}
\usepackage{pbox}
\usepackage{comment}
\usepackage{ifthen}
\usepackage{color}
\usepackage{colortbl,xcolor}
\usepackage{upgreek}
\usepackage{upgreek}
\usepackage{todonotes}
\usepackage{breakurl}

\usepackage{tikz}
\usetikzlibrary{arrows}

\usepackage[linesnumbered,ruled,vlined]{algorithm2e}
\SetKwInput{KwInput}{Input}
\SetKwInput{KwOutput}{Output}
\SetKwProg{myproc}{Procedure}{}{}

\colorlet{punct}{red!60!black}
\definecolor{background}{HTML}{EEEEEE}
\definecolor{delim}{RGB}{20,105,176}
\colorlet{numb}{magenta!60!black}

\definecolor{Gray}{gray}{0.9}
\definecolor{LightCyan}{rgb}{0.88,1,1}

\specialcomment{notes}{\begingroup \color{blue}} { \endgroup}

\newcolumntype{!}{>{\global\let\currentrowstyle\relax}}
\newcolumntype{^}{>{\currentrowstyle}}

\newcommand{\si}{\begin{enumerate}}

\newcommand{\ei}{\end{enumerate}}

\makeatletter
\let\oldfootnote\footnote
\def\footnote{\@ifstar\footnote@star\footnote@nostar}
\def\footnote@star#1{{\let\thefootnote\relax\footnotetext{#1}}}
\def\footnote@nostar{\oldfootnote}
\makeatother

\makeatletter
\renewcommand\footnotesize{%
   \@setfontsize\footnotesize\@ixpt{5}%
   \abovedisplayskip 8\p@ \@plus2\p@ \@minus4\p@
   \abovedisplayshortskip \z@ \@plus\p@
   \belowdisplayshortskip 4\p@ \@plus2\p@ \@minus2\p@
   \def\@listi{\leftmargin\leftmargini
               \topsep 4\p@ \@plus2\p@ \@minus2\p@
               \parsep 2\p@ \@plus\p@ \@minus\p@
               \itemsep \parsep}%
   \belowdisplayskip \abovedisplayskip
}
\makeatother

\lstdefinelanguage{json}{
    basicstyle=\small\ttfamily,
    numbers=left,
    numbers=none,
    stepnumber=1,
    numbersep=8pt,
    showstringspaces=false,
    breaklines=true,
    frame=lines,
    backgroundcolor=\color{background},
    literate=
     *{0}{{{\color{numb}0}}}{1}
      {1}{{{\color{numb}1}}}{1}
      {2}{{{\color{numb}2}}}{1}
      {3}{{{\color{numb}3}}}{1}
      {4}{{{\color{numb}4}}}{1}
      {5}{{{\color{numb}5}}}{1}
      {6}{{{\color{numb}6}}}{1}
      {7}{{{\color{numb}7}}}{1}
      {8}{{{\color{numb}8}}}{1}
      {9}{{{\color{numb}9}}}{1}
      {:}{{{\color{punct}{:}}}}{1}
      {,}{{{\color{punct}{,}}}}{1}
      {\{}{{{\color{delim}{\{}}}}{1}
      {\}}{{{\color{delim}{\}}}}}{1}
      {[}{{{\color{delim}{[}}}}{1}
      {]}{{{\color{delim}{]}}}}{1},
}

\DeclareGraphicsExtensions{.eps,.png}
\graphicspath{{figure/}{figure/eps/}{figure/png/}}

\makeatletter
\def\@copyrightspace{\relax}
\makeatother

\lstdefinelanguage{json}{
    basicstyle=\small\ttfamily,
    numbers=left,
    numbers=none,
    stepnumber=1,
    numbersep=8pt,
    showstringspaces=false,
    breaklines=true,
    frame=lines,
    backgroundcolor=\color{background},
    literate=
     *{0}{{{\color{numb}0}}}{1}
      {1}{{{\color{numb}1}}}{1}
      {2}{{{\color{numb}2}}}{1}
      {3}{{{\color{numb}3}}}{1}
      {4}{{{\color{numb}4}}}{1}
      {5}{{{\color{numb}5}}}{1}
      {6}{{{\color{numb}6}}}{1}
      {7}{{{\color{numb}7}}}{1}
      {8}{{{\color{numb}8}}}{1}
      {9}{{{\color{numb}9}}}{1}
      {:}{{{\color{punct}{:}}}}{1}
      {,}{{{\color{punct}{,}}}}{1}
      {\{}{{{\color{delim}{\{}}}}{1}
      {\}}{{{\color{delim}{\}}}}}{1}
      {[}{{{\color{delim}{[}}}}{1}
      {]}{{{\color{delim}{]}}}}{1},
}

\title{Lithium NLP: A System for Rich Information Extraction from Noisy User Generated Text on Social Media}

\author{Preeti Bhargava \and Nemanja Spasojevic \and Guoning Hu \\
Lithium Technologies $\vert$ Klout \\
San Francisco, CA\\
  {\tt {{preeti.bhargava, nemanja.spasojevic, guoning.hu}@lithium.com}}}

\begin{document}

\maketitle

\begin{abstract}

In this paper, we describe the Lithium Natural Language Processing (NLP) system - a resource-constrained, high-throughput and language-agnostic system for  information extraction from noisy user generated text on social media. Lithium NLP extracts a rich set of information including entities, topics, hashtags and sentiment from text. We discuss several real world applications of the system currently incorporated in Lithium products. We also compare our system with existing commercial and academic NLP systems in terms of performance, information extracted and languages supported. We show that Lithium NLP is at par with and in some cases, outperforms state-of-the-art commercial NLP systems.
\end{abstract}

%

\section{Introduction}
\label{section:introduction}

\begin{figure*}[t]
  \centering
  \begin{minipage}[b]{0.45\textwidth}
    \includegraphics[width=\textwidth]{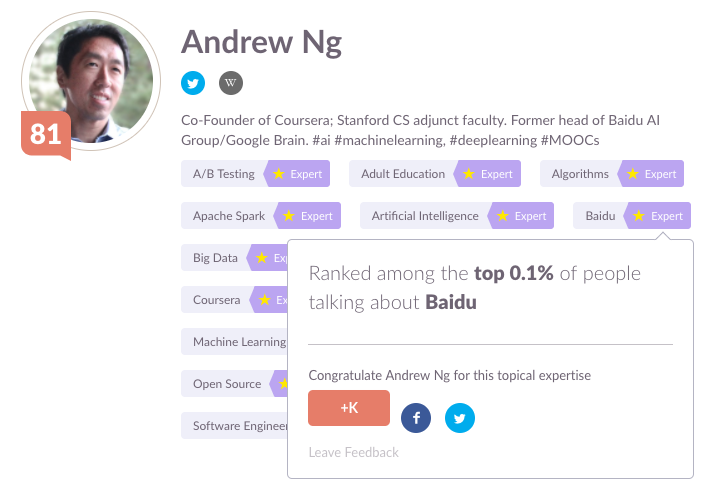}
    \caption{A user's inferred expertise topics}
    \label{fig:expertise_score}
  \end{minipage}
  \qquad
  \begin{minipage}[b]{0.45\textwidth}
    \includegraphics[width=\textwidth]{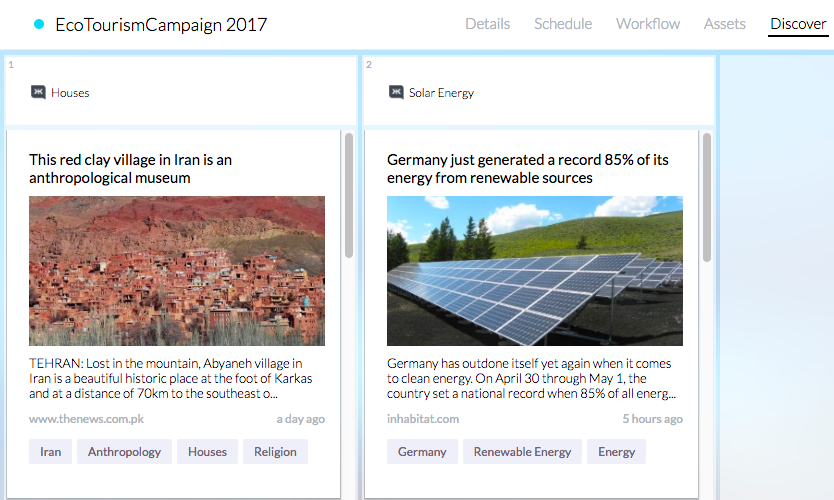}
    \caption{Content Personalization}
    \label{fig:content_discovery}
  \end{minipage}
\end{figure*}

Social media has become one of the major means for communication and content production. As a result, industrial systems that possess the capability to process rich user generated content from social media platform have several real-world applications. Furthermore, due to the content style, size and heterogeneity of information (e.g. text, emoticons, hashtags etc.) available on social media, novel NLP techniques and systems that are designed specifically for such content and can potentially integrate or learn information from different sources are highly useful and applicable. 

However, NLP on social media data can be significantly complex and challenging due to several reasons:
\begin{itemize}
\item \textbf{Noisy unnormalized data} - Social media data is much more informal than traditional text and less consistent in language in terms of style, tone etc. It involves heavy usage of slang, jargons, emoticons, or abbreviations which usually do not follow formal grammatical rules. Hence, novel NLP techniques need to be developed for such content.
\item \textbf{Multi-lingual content} - Social media data poses an additional challenge to NLP practitioners because the user generated content on them is often multi-lingual. Hence, any NLP system processing real world data from the web should be able to support multiple languages in order to be practical and applicable. 
\item \textbf{Large scale datasets} - State-of-the-art NLP systems should be able to work on large scale datasets such as social media data, often involving millions of documents. Moreover, these systems need to have low resource consumption in order to scale to such datasets in a finite amount of time. In addition, in order to be applicable and practical, they should be able to run on off-the-shelf commodity machines. 
\item \textbf{Rich set of information} - In order to be cost-efficient, state-of-the-art NLP systems need to be exhaustive in terms of information extracted\footnote{\url{https://en.wikipedia.org/wiki/Information_extraction}} from social media text. This includes extracting entities of different types (such as professional titles, sports, activities etc.) in addition to just named entities (such as persons, organizations, locations etc.), inferring fine-grained and coarse-grained subject matter topics (sports, politics, healthcare, basketball), text sentiment, hashtags, emoticons etc. 
\end{itemize}

In this paper, we present the Lithium NLP\footnote{A screencast video demonstrating the system is available at \url{https://youtu.be/U-o6Efh6TZc}} system which addresses these challenges. It is a resource-constrained, high-throughput and language-agnostic system for information extraction from noisy user generated text such as that available on social media. It is capable of extracting a rich set of information including entities, topics, hashtags and sentiment. Lithium NLP currently supports multiple languages including Arabic, English, French, German, Italian and Spanish. It supports large scale data from several social media platforms such as Twitter, Facebook, Linkedin, etc. by processing about $500M$ new social media messages, and $0.5M$ socially relevant URLs shared daily. Since it employs statistical NLP techniques, it uses the large scale of the data to help overcome the noisiness.

Lithium NLP is currently incorporated in several Lithium products. It enables consumer products like Klout\footnote{\url{https://klout.com}} - a platform which integrates users' data from multiple social networks such as Twitter, Facebook, Instagram, Linkedin, GooglePlus, Youtube, and Foursquare, in order to measure their online social influence via the \emph{Klout Score\footnote{\url{https://klout.com/corp/score}}} \cite{rao2015kloutScore}. On Klout, it is used to model users' topics of interest \cite{nemanja-lasta} and expertise \cite{Spasojevic2016:experts} by building their topical profiles. Figure \ref{fig:expertise_score} shows an example of a user's topics of expertise, as inferred on Klout. Currently, we build topical profiles for more than $600M$ users. These profiles are further used to recommend personalized content to these users by matching their topics of interest or expertise with content topics as this leads to better user engagement. 
An example of content personalization is shown in Figure \ref{fig:content_discovery}. The user scores and topics are also available via the GNIP PowerTrack API\footnote{\url{http://support.gnip.com/enrichments/klout.html}}.

Lithium NLP also enables enterprise products such as Lithium's social media management tools\footnote{\url{https://www.lithium.com/products/social-media-management/}} - Lithium Reach and Lithium Response. It is used to analyze $20+M$ new daily engagements across Lithium's 400+ communities\footnote{\url{https://www.lithium.com/products/online-communities/}}. In the past, a version of Lithium NLP had been used to enable user targeting applications such as Klout Perks\footnote{\url{https://goo.gl/vtZDqE\#Klout_Perks}} (influencer reward platform), Cinch\footnote{\url{https://goo.gl/CLcx9p\#Cinch}} (Q\&A app), and Who-To-Follow recommendations. These involved selecting a group of users for targeting based on given topics and other filtering criteria.





\section{Knowledge Base}
\label{section:knowledgebase}
Our Knowledge Base (KB) consists of about 1 million Freebase machine ids for entities 
that were chosen from a subset of  all Freebase entities that map to Wikipedia entities.
We  prefer to use Freebase rather than Wikipedia as our KB since in Freebase, the same id represents a unique entity across multiple languages.
Due to limited resources and usefulness of the entities, our KB contains approximately 1 million most important entities from among all the Freebase entities. 
This gives us a good balance between coverage and relevance of entities for processing common social media text.
Section \ref{subsection:resourcegeneration} explains how entity importance is calculated, which enables us to rank the top 1 million Freebase entities.

In addition to the KB entities, we also employ two special entities: \textbf{NIL} and \textbf{MISC}.
\textbf{NIL} entity indicates that there is no entity associated with the mention, eg. mention `the' within the sentence may link to entity \textbf{NIL}. This entity is useful especially when it comes to dealing with stop words and false positives. \textbf{MISC} indicates that the mention links to an entity which is outside the selected entity set in our KB. 

%

\section{System Overview}
\label{section:systemarch}
Figure \ref{fig:papyrus} shows a high level overview of the Lithium NLP system. It has two phases:

\begin{figure*}[t]
\begin{minipage}[b]{1\linewidth} 
\centering
\includegraphics[width=\textwidth]{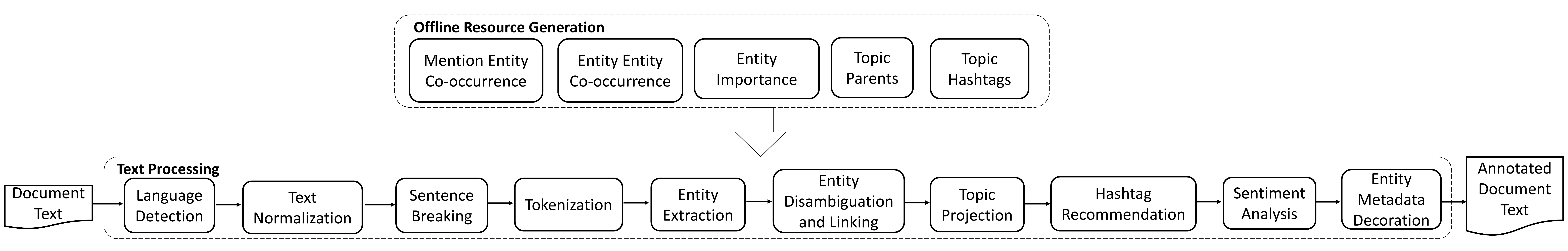}
\caption{Overview of the Lithium NLP pipeline}
\label{fig:papyrus}
\end{minipage}

\begin{minipage}[b]{1\linewidth} 
\centering
\includegraphics[width=\textwidth]{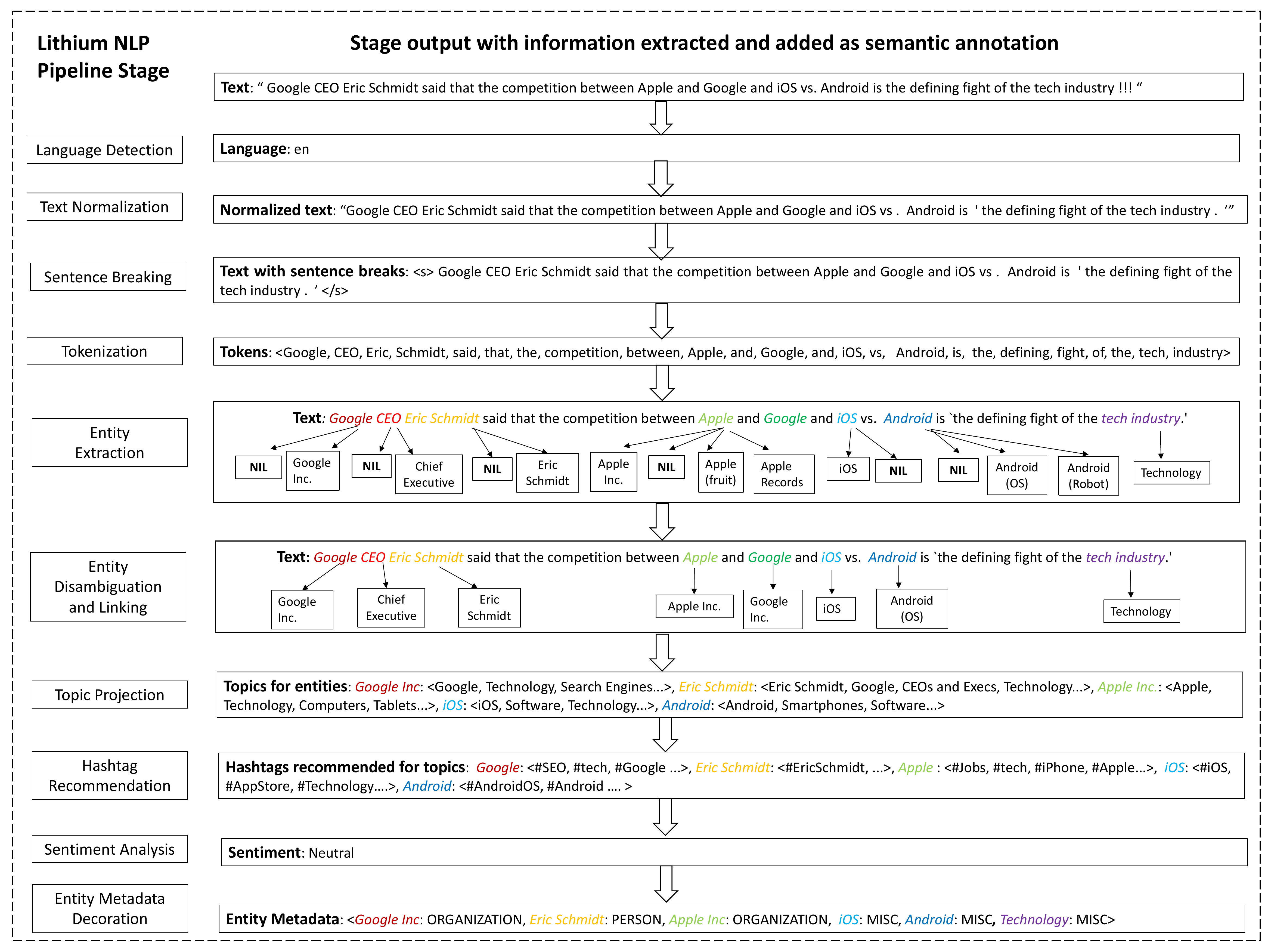}
\caption{An example demonstrating the information extracted and added as semantic annotation at each stage of the Lithium NLP pipeline (best viewed in color)}
\label{fig:papyrusdemo}
\end{minipage}
\end{figure*}

\subsection{Offline Resource Generation \label{subsection:resourcegeneration}}

In this phase, we generate several dictionaries that capture language models, probabilities and relations across entities and topics, by leveraging various multi-lingual data sources. Some of these dictionaries are derived using our DAWT\footnote{\url{https://github.com/klout/opendata/tree/master/wiki_annotation}} data set \cite{Spasojevic:dawt} that consists of densely annotated wikipedia pages across multiple languages. It is 4.8 times denser than Wikipedia and is designed to be exhaustive across several domains. 

The dictionaries generated from the DAWT dataset are:

\begin{itemize}
  \item \textbf{Mention-Entity Co-occurrence} - This dictionary captures the prior probability that a mention $M_{i}$ refers to an entity $E_{j}$ (including \textbf{NIL} and \textbf{MISC}) within the DAWT dataset
and is equivalent to the cooccurrence probability of the mention and the entity:
 
 \small
$${{count (M_{i} \rightarrow E_{j})} \over {count (M_{i})}}$$
\normalsize

For instance, mention \emph{Michael Jordan} can link to \textbf{Michael Jordan (Professor)} or \textbf{Michael Jordan (Basketball player)} with different prior probabilities. 
Moreover, we generate a separate dictionary for each language. 

  \item \textbf{Entity-Entity Co-occurrence} - This dictionary captures co-occurrence frequencies among entities by counting all the entities that simultaneously appear within a sliding window of 50 tokens. 
Moreover, this data is accumulated across all languages and is language independent in order to capture better relations and create a smaller memory footprint when supporting additional languages. Also, for each entity, we consider only the top 30 co-occurring entities which have at least 10 or more co-occurrences across all supported languages. For instance, entity \textbf{Michael Jordan (Basketball player)} co-occurs with entities \textbf{Basketball}, \textbf{NBA} etc. while entity \textbf{Michael Jordan (Professor)} co-occurs with entities \textbf{Machine Learning}, \textbf{Artificial Intelligence}, \textbf{UC Berkeley} etc.
\end{itemize}

We also generate additional dictionaries:
\begin{itemize}
  \item \textbf{Entity Importance} - The entity importance score \cite{Bhattacharyya-importance} is derived as a global score identifying how important an extracted entity is for a casual observer. This score is calculated using linear regression with features capturing popularity within Wikipedia links, and importance of the entity within Freebase. We used signals such as Wiki page rank, Wiki and Freebase incoming and outgoing links, and type descriptors within our KB etc.

  \item \textbf{Topic Parents} -  This dictionary contains the parent topics for each topic in the Klout Topic Ontology \footnote{\url{https://github.com/klout/opendata/tree/master/klout_topic_ontology}} (KTO) - a manually curated ontology built to capture social media users' interests and expertise scores, in different topics, across multiple social networks. As of April 2017, it consists of roughly 8,030 topic nodes and 13,441 edges encoding hierarchical relationships among them.  

  \item \textbf{Topic Hashtags} -  This dictionary contains hashtags recommended for topics in KTO. We determine the hashtags via co-occurrence counts of topics and hashtags, importance, recency and popularity of hashtags as well popularity of topics.
  

\end{itemize}

\subsection{Text Processing}

In the Lithium NLP system, an input text document is stored as a Protocol Buffers\footnote{\url{https://developers.google.com/protocol-buffers/}} message. The Text Processing phase of the system processes the input text document through several stages and the information (entities, topics etc.) extracted at every stage is added as a semantic annotation to the text. 
Not all annotations are added to a document, the Lithium NLP API (explained in Section \ref{subsection:api}) allows a client application to select specific annotations. However, certain annotations such as language and tokens are prerequisites for later stages. 

The Text Processing pipeline stages are:

\begin{itemize}
\item \textbf{Language Detection} - This stage detects the language of the input document using an open source language detector\footnote{\url{https://github.com/shuyo/language-detection}}. This detector employs a naive Bayesian filter which uses character, spellings and script as features to classify language and estimate its probability. It has a precision of $~99\%$ for 49 languages. 
\item \textbf{Text Normalization} - This stage normalizes the text by escaping unescaped characters and replacing special characters (e.g. diacritical marks) based on the detected language. It replaces non-ASCII punctuations and hyphens with spaces, multiple spaces with single space, converts accents to regular characters etc.

\item \textbf{Sentence Breaking} - This stage breaks the normalized text into sentences using Java Text API\footnote{\url{https://docs.oracle.com/javase/7/docs/api/java/text/BreakIterator.html}}. It can distinguish sentence breakers from other marks, such as periods within numbers and abbreviations, according to the detected language.

\item \textbf{Tokenization} - This stage converts each sentence into a sequence of tokens via the Lucene Standard Tokenizer\footnote{\url{http://lucene.apache.org/core/4_5_0/analyzers-common/org/apache/lucene/analysis/standard/StandardTokenizer.html}} for all languages and the Lucene Smart Chinese Analyzer\footnote{\url{https://lucene.apache.org/core/4_5_0/analyzers-smartcn/org/apache/lucene/analysis/cn/smart/SmartChineseAnalyzer.html}} for Chinese.

\item \textbf{Entity Extraction} - This stage extracts mentions in each sentence using the Mention Entity Co-occurrence dictionary generated offline (Section \ref{subsection:resourcegeneration}).
%
A mention may contain a single token or several consecutive tokens, but a token can belong to at most one mention. 

To make this task computationally efficient, we apply a simple greedy strategy that analyzes windows of \emph{n}-grams (n $\in$ [1,6]) and extracts the longest mention found in each window. For each extracted mention, we generate multiple candidate entities. For instance, mention \emph{Android} can link to candidate entities \textbf{Android (OS)} or \textbf{Android (Robot)}.

\item \textbf{Entity Disambiguation and Linking (EDL)} - This stage disambiguates and links an entity mention to the correct candidate entity in our KB \cite{Bhargava:edl}. 
It uses several features obtained from the dictionaries generated offline (Section \ref{subsection:resourcegeneration}). These include context-independent features, such as mention-entity co-occurrence, mention-entity Jaccard similarity and entity importance, and context-dependent features such as entity entity co-occurrence and entity topic semantic similarity. It employs machine learning models, such as decision trees and logistic regression, generated using these features to correctly disambiguate a mention and link to the corresponding entity. This stage has a precision of 63\%, recall of 87\% and an F-score of 73\% when tested on an in-house dataset.

\item \textbf{Topic Projection} -  In this stage, we associate each entity in our KB to upto 10 most relevant topics in KTO. For instance, entity \textbf{Android (OS)} will be associated with the topics such as \emph{Smartphones}, \emph{Software} etc. 

We use a weighted ensemble of several semi-supervised models that employ entity co-occurrences, GloVe \cite{glove2014} word vectors, Freebase hierarchical relationships and Wikipedia in order to propagate topic labels. A complete description of this algorithm is beyond the scope of this paper.

\item \textbf{Hashtag Recommendation} - In this stage, we annotate the text with hashtags recommended based on the topics associated with the text in Topic Projection. This uses the Topic Hashtags dictionary generated offline (Section \ref{subsection:resourcegeneration})

\item \textbf{Sentiment Analysis} - In this stage, we determine the sentiment of the text (positive, negative or neutral) via lexicons and term counting with negation handling \cite{Spasojevic:actionability}. For this, we used several lexicons of positive and negative words (including SentiWordNet \cite{baccianella2010sentiwordnet, esuli2007sentiwordnet} and AFINN \cite{nielsen2011new}) as well as emoticons. We compute the sentiment score as

\small
$$
{{W_{Pos} - W_{Neg}} \over {\textrm{Log(Total \# of words in text) }+ \epsilon}}
$$
\normalsize

where $W_{Pos}$ is the weighted strength of positive words and emoticons, $W_{Neg}$ is the weighted strength of negative words and emoticons in the text and $\epsilon$ is a smoothing constant. If the score is positive and above a certain threshold, the text is classified as `Positive'. If it is below a certain threshold, the text is classified as `Negative'. If it lies within the boundary between `Positive' and `Negative' classes, the text is classified as `Neutral'. 

To handle negations, we use a \emph{lookback window}. Every time, we encounter a word from our sentiment lexicons, we look back at a window of size 3 to see if any negation words precede it and negate the weight of the sentiment word. Overall, this stage has a precision of 47\%, recall of 48\% and an F-score of 46\% when tested on an in-house dataset.

\item \textbf{Entity Metadata Decoration} - In this stage, we add the entity metadata such as its type (Person, Organization, Location, Film, Event, Book) and Location (Population, Time Zone, Latitude/Longitude).

\end{itemize}

Figure \ref{fig:papyrusdemo} demonstrates how the Lithium NLP pipeline processes a sample text ``\emph{Google CEO Eric Schmidt said that the competition between Apple and Google and iOS vs. Android is `the defining fight of the tech industry'.}" and adds the annotations at every stage.

\subsection{REST API}\label{subsection:api}

The Lithium NLP system provides a REST API via which client applications can send a text document as request and receive the annotated text as JSON response. A snippet of an annotated response (which is in our text proto format\footnote{\url{https://github.com/klout/opendata/blob/master/wiki_annotation/Text.proto}}) received through the API is shown in Listing \ref{jsonresponse}. Note that the disambiguated entities are also linked to their Freebase ids and Wikipedia links. \\ \\ 

\begin{lstlisting}[language=json,basicstyle=\scriptsize,firstnumber=1,breaklines=true,caption=JSON of annotated text summary,label=jsonresponse]
{
  "text": "Vlade Divac Serbian NBA player used to play for LA Lakers.",
  "language": "en",
  "annotation_summary": [{
      "type": "ENTITY",
      "annotation_identifier": [{
          "id_str": "01vpr3",
          "id_url": "https://en.wikipedia.org/wiki/Vlade_Divac",
          "score": 0.9456,
          "type": "PERSON"
        }, {
          "id_str": "05jvx",
          "id_url": "https://en.wikipedia.org/wiki/NBA",
          "score": 0.8496,
          "type": "ORGANIZATION"
        }, ...
      }]
  },
  {
    "type": "KLOUT_TOPIC",
    "annotation_identifier": [{
      "id_str": "6467710261455026125",
      "id_readable": "nba",
      "score": 0.7582
    }, {
      "id_str": "8311852403596174326",
      "id_readable": "los-angeles-lakers",
      "score": 0.66974
    }, {
      "id_str": "8582816108322807207",
      "id_readable": "basketball",
      "score": 0.5445
    }, ...]
  },
  {
    "type": "HASHTAG",
    "annotation_identifier": [{
      "id_str": "NBA",
      "score": 54285.7515
    }, {
      "id_str": "NBAPlayoffs",
      "score": 28685.6006
    }, ...]
  }],
   "sentiment": 0.0
}
\end{lstlisting}

\subsection{Performance}\label{subsection:performance}

\begin{figure}[t]
  \centering
  \includegraphics[width=0.47\textwidth,height=70mm]{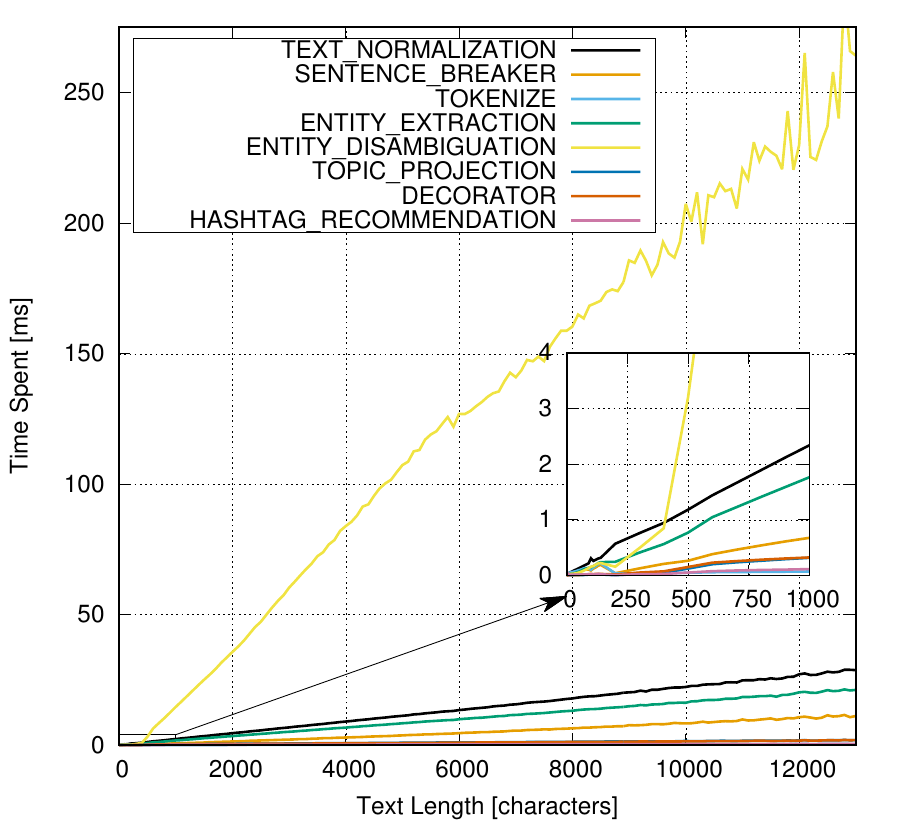}
  \caption{Lithium NLP performance per processing stage (best viewed in color)}
  \label{fig:stage_performance}
   \vspace{-0.1in}
\end{figure}

Figure \ref{fig:stage_performance} shows the computational performance per processing stage of the Lithium NLP system.
The overall processing speed is about 22ms per 1kb of text. As shown, the time taken by the system is a linear function
of text size. The EDL stage takes about $80\%$ of the processing time.



\section{Comparison with existing NLP systems}
\label{section:relatedwork}
\begin{figure}[t]
  \centering
  \includegraphics[width=0.47\textwidth]{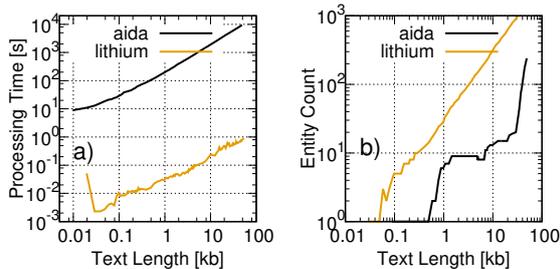}
  \caption{AIDA vs. Lithium NLP  Comparison on \textbf{a)} Text processing runtime \textbf{b)} Extracted entity count  (best viewed in color)}
  \label{fig:aida_vs_lithium}
\end{figure}

\begin{table*}[t]
\begin{minipage}[b]{1\linewidth} 
\centering
\resizebox{\textwidth}{!}{%
\begin{tabular}{|c|c|c|c|c|c|c|c|}
\hline
 & \textbf{Lithium NLP} & \textbf{Google NL} & \textbf{Open Calais} & \textbf{Alchemy API} & \textbf{Stanford CoreNLP} & \textbf{Ambiverse} & \textbf{Twitter NLP}\\
  \hline
  \textbf{Named Entities} & X & X & X & X & X & X & X \\
    \hline
  \textbf{Other Entities} & X & X & X & X & X & &\\
\hline
  \textbf{Topics (fine-grained)} & X &  &  &  &  &  &\\
  \hline
  \textbf{Topics (coarse-grained)} & X	&  & X & X &  & &  \\
\hline
  \textbf{Hashtags} & X &  &  &  &  &  &\\
\hline
  \textbf{Document Sentiment} & X &  &  & X & X & &\\
\hline   
  \textbf{Entity level Sentiment} &  & X &  & X & & &\\
  \hline
  \textbf{Entity types} & X & X & X & X &X &X & X\\
  \hline
    \textbf{Relationships} &  &  & X & X &X & &\\
  \hline
      \textbf{Events} &  &  &  &  & & & X\\
  \hline
\end{tabular}}
\caption{Comparison of information extracted by Lithium NLP with existing NLP systems}
\label{table:CapabilitiesComparison}
\end{minipage}

\begin{minipage}[b]{1\linewidth}
\centering
\resizebox{\textwidth}{!}{%
\begin{tabular}{|c|c|c|c|c|c|c|c|}
\hline
 & \textbf{Lithium NLP} & \textbf{Google NL} & \textbf{Open Calais}  & \textbf{Alchemy API}  & \textbf{Stanford CoreNLP} & \textbf{Ambiverse} & \textbf{Twitter NLP}\\
\hline
 \textbf{Supported}  & Arabic, English, & Chinese, English, French, & English,  & English, French, German,  & Arabic, Chinese,  & English, German, & English\\ 
 \textbf{Languages} &  French, German,  &  German, Italian, Japanese,  & French, &  Italian, Portuguese, Russian,  & English, French,  & Spanish,  & \\ 
  &  Italian, Spanish & Korean, Portugese, Spanish & Spanish &  Spanish, Swedish  & German, Spanish & Chinese & \\ 
  \hline
\end{tabular}}
\caption{Comparison of languages supported by Lithium NLP with existing NLP systems}
\label{table:LanguagesComparison}
\end{minipage} 
\end{table*}

Currently, due to limited resources at our end and also due to inherent differences in the Knowledge Base (Freebase vs Wikipedia and others), test dataset, and types of information extracted (entities, topics, hashtags etc.), a direct comparison of the Lithium NLP system's performance (in terms of precision, recall and f-score) with existing academic and commercial systems such as Google Cloud NL API\footnote{\url{https://cloud.google.com/natural-language/}}, Open Calais\footnote{\url{http://www.opencalais.com/opencalais-demo/}}, Alchemy API\footnote{\url{https://alchemy-language-demo.mybluemix.net/}}, Stanford CoreNLP\footnote{\url{http://corenlp.run/}} \cite{manning2014stanford}, Ambiverse/AIDA\footnote{\url{https://www.ambiverse.com/}} \cite{nguyen2014aida} and Twitter NLP\footnote{\url{https://github.com/aritter/twitter_nlp}} \cite{ritter2011named,Ritter12} is not possible. Hence, we compare our system with some of them on a different set of metrics.

\subsection{Comparison on runtime and entity density}

We compare the runtime of Lithium NLP and AIDA across various text sizes.
As shown in Figure \ref{fig:aida_vs_lithium}, Lithium NLP is on an average 40,000 times faster than AIDA whose slow runtime can be attributed mainly to Stanford NER. In addition to speed, we also compare the number of entities extracted per kb of text. As shown, Lithium NLP extracts about $2.8$ times more entities than AIDA.

\subsection{Comparison on information extracted}

Table \ref{table:CapabilitiesComparison} compares the types of information extracted by Lithium NLP system with existing systems. In this comparison, we explicitly differentiate between named entities (Person, Location etc.) and other entity types (Sports, Activities) as well as fine-grained topics (Basketball) and coarse-grained topics (Sports) to demonstrate the rich set of information extracted by Lithium NLP. As evident, most other systems do not provide the rich set of semantic annotations that Lithium NLP provides. A majority of the systems focus on recognizing named entities and types with only a few focusing on sentiment and coarse-grained topics as well. In contrast, Lithium NLP extracts, disambiguates and links named and other entities, extracts subject matter topics, recommends hashtags and  also infers the sentiment of the text.

\subsection{Comparison on languages}

Table \ref{table:LanguagesComparison} compares the languages supported by the Lithium NLP system with existing systems. As evident, Lithium supports 6 different languages which is at par and in some cases, more than existing systems.

\section{Conclusion and Future Work}
\label{section:conclusion}
%

In this paper, we described the Lithium NLP system - a resource-constrained, high-throughput and language-agnostic system for information extraction from noisy user generated text on social media. Lithium NLP extracts a rich set of information including entities, topics, hashtags and sentiment from text. We discussed several real world applications of the system currently incorporated in Lithium products. We also compared our system with existing commercial and academic NLP systems in terms of performance, information extracted and languages supported. We showed that Lithium NLP is at par with and in some cases, outperforms state-of-the-art commercial NLP systems.

In future, we plan to extend the capabilities of Lithium NLP to include entity level sentiment as well. We also hope to collaborate actively with academia and open up the Lithium NLP API to academic institutions.

\section*{Acknowledgements}
\label{section:acknowledgements}
The authors would like to thank Prantik Bhattacharya, Adithya Rao and Sarah Ellinger for their contributions to the Lithium NLP system. They would also like to thank Mike Ottinger and Armin Broubakarian for their help with building the Lithium NLP UI and demo.


\bibliographystyle{emnlp_natbib}
\bibliography{bibliography}

\end{document}